\documentclass{article}

\usepackage[utf8]{inputenc} 
\usepackage[T1]{fontenc}    
\usepackage{hyperref}       
\usepackage{url}            
\usepackage{booktabs}       
\usepackage{amsmath}
\usepackage{amsfonts}       
\usepackage{nicefrac}       
\usepackage{microtype}      
\usepackage{tabularx}

\usepackage{graphicx}   
\usepackage{subfigure} 

\usepackage{color}	
\usepackage[colorinlistoftodos]{todonotes}

\usepackage{ulem}

\usepackage{geometry}
\title{Evenly Cascaded Convolutional Networks \\
}

\author{Chengxi Ye, Chinmaya Devaraj, Michael Maynord, \\Cornelia Ferm\"{u}ller, Yiannis Aloimonos
\\
University of Maryland Institute for
Advanced Computer Studies, 
\\
\{cxy, chinmayd, maynord, fer, yiannis\}@umiacs.umd.edu
}

\begin{document}

\maketitle

\begin{abstract}

We introduce Evenly Cascaded convolutional Network (ECN), a neural network taking inspiration from the cascade algorithm of wavelet analysis. ECN employs two feature streams - a low-level and high-level steam. At each layer these streams interact, such that low-level features are modulated using advanced perspectives from the high-level stream. ECN is evenly structured through resizing feature map dimensions by a consistent ratio, which removes the burden of ad-hoc specification of feature map dimensions. ECN produces easily interpretable features maps, a result whose intuition can be understood in the context of scale-space theory. We demonstrate that ECN's design facilitates the training process through providing easily trainable shortcuts. We report new state-of-the-art results for small networks, without the need for additional treatment such as pruning or compression - a consequence of ECN's simple structure and direct training. A 6-layered ECN design with under 500k parameters achieves 95.24\% and 78.99\% accuracy on CIFAR-10 and CIFAR-100 datasets, respectively, outperforming the current state-of-the-art on small parameter networks, and a 3 million parameter ECN produces results competitive to the state-of-the-art.

\end{abstract}


\begin{figure}
\centering
\includegraphics[width=3 in]{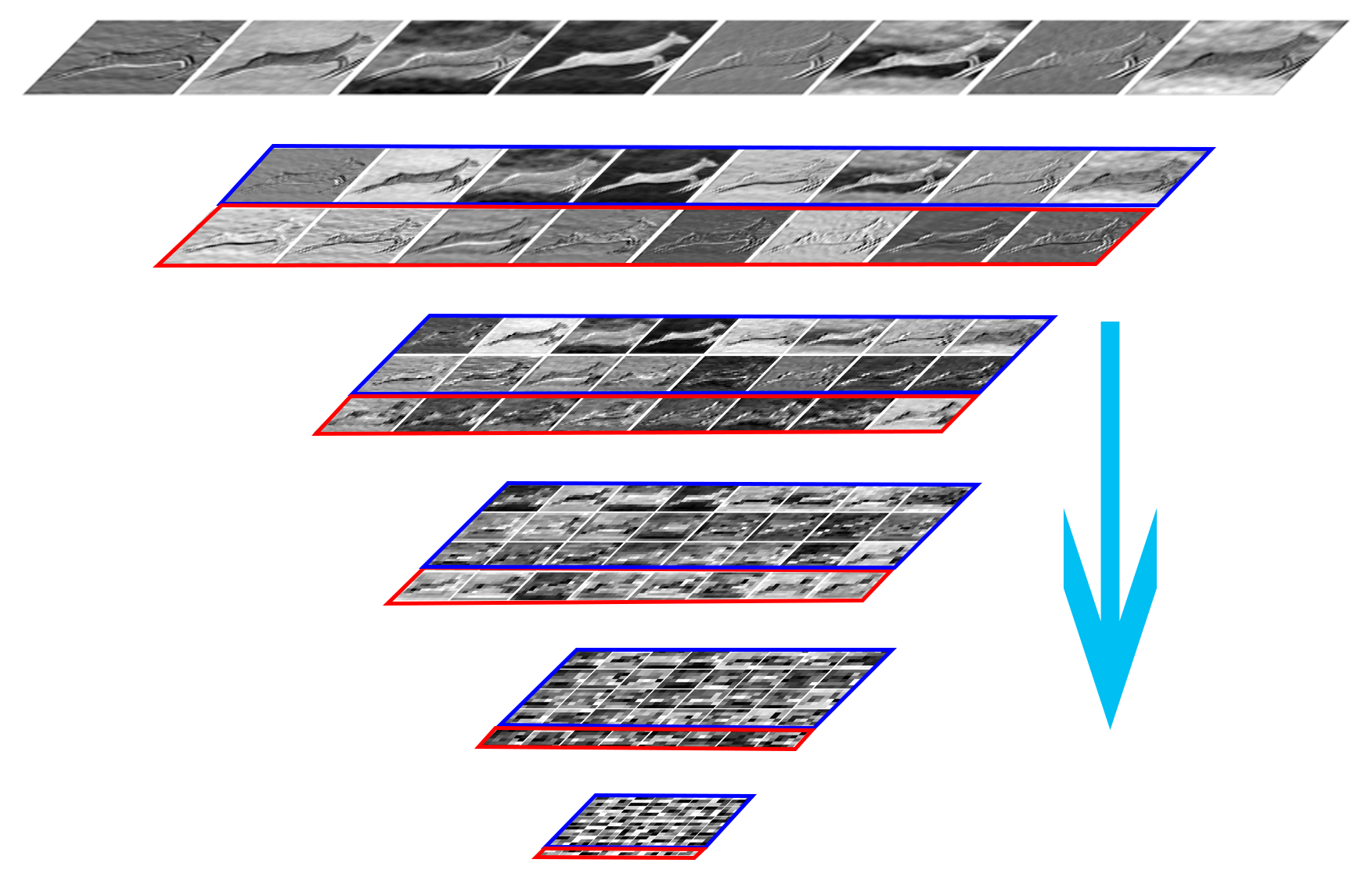}
\caption{Our evenly cascaded design with $6$ cascading layers using a scaling rate of $0.75$. The growth rate (the number of high level feature channels generated in each subsequent layer) is set to $8$. From the second layer on, the low level feature channels are framed in blue boxes, the newly generated high level feature channels are framed in red boxes.}
\label{fig:CasNet}
\end{figure}

\section{Introduction} \label{sec:introduction}

How do humans come to understand their world? Consider the learning of mathematics - we learn basic mathematics before proceeding to advanced mathematics, the understanding of basic math serving as the foundation on which to develop an understanding of advanced math. Once an advanced understanding of math is acquired, it in turn is used to better understand basic math, reinforcing an understanding of mathematics as a whole. Similarly in perception - higher levels of abstraction within our perceptual hierarchies are built upon the lower levels of abstraction. Additionally, using higher level interpretations to reinforce lower level interpretations can provide benefits to perception as a whole, in the same way that using a more expansive understanding of mathematics to understand elementary mathematics reinforces an understanding of mathematics as a whole.

Disregarding the interplay between levels of abstraction in perception, the recent trend in deep neural networks is to simply make networks deeper. 
Since networks have become so deep, researchers have developed a now standard design that divides network layers into blocks of layers~\cite{krizhevsky2012imagenet,VGG,he2016deep,Huang2016Densely}. Each block consists of layers of transforms that produce feature maps of the same shape. Cross-block transforms recombine the previous features and incorporate strided convolutions or pooling operations to reshape the feature maps. This strategy is an extension of the traditional design of the multilayer perceptron or fully-connected network~\cite{Rumelhart:1986:LIR:104279.104293}. The introduction of blocks has allowed a better organization of the evolving layers of abstraction within those networks. Overall, more high level features are abstracted through these transforms (Fig.~\ref{fig:CasNet},~\ref{fig:Multilevel}). However, the unstructured recombination of features in existing networks has made the investigation of deep neural networks nontrivial~\cite{DBLP:journals/corr/ZeilerF13,UnderstandingCNN_Mallat}.

Aside from deviating from basic intuitions about learning and perception, the now standard design has other apparent shortcomings. Since strided convolution or pooling are used to resize the feature maps by integer scales, researchers are compelled to create ad hoc network shapes, ones which are subject to awkward constraints of integer pooling and strides. Some networks assign more layers in later stages, where the feature maps are very small~\cite{he2016deep}. Some other networks are wider but have fewer layers~\cite{WideResNet}. The specification of feature map dimensions or distribution of computational resources is highly engineered and uneven in these designs. It is difficult to determine which designs outperform others, given that the overall shape is different. To make things worse, when the task changes, the network shape needs to be handcrafted again.  
 
Arguably the biggest shortcoming of existing designs is that they afford little intuition on the training process, and consequently make deep networks notoriously hard to train. Researchers have expended significant effort in developing better methods to train these networks.

In this work we introduce an architecture more closely in line with intuitions of biological learning and perception: Evenly Cascaded convolutional Network (ECN). ECN is 1) easier than existing architectures to adapt to new tasks, 2) produces internal representations which are more humanly interpretable, 3) performs robustly as parameter count is restricted, and 4) produces competitive performance when compared to other state-of-the-art methods.

\begin{figure}
\centering
\subfigure[]{\includegraphics[height=.6 in]{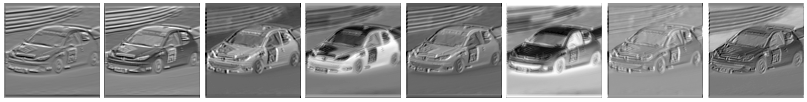}}
\subfigure[]{\includegraphics[height=.8 in]{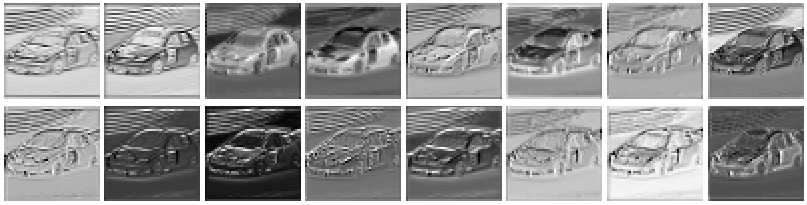}}
\subfigure[]{\includegraphics[height=.8 in]{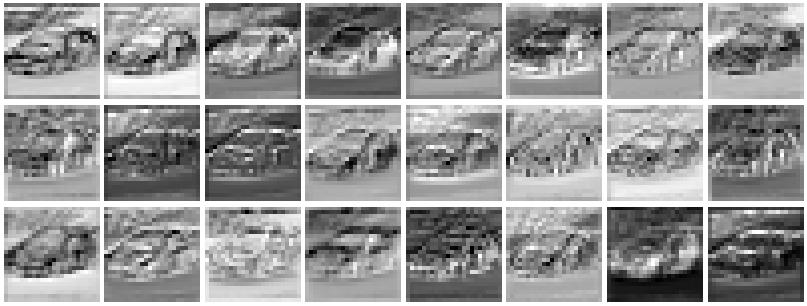}}
\subfigure[]{\includegraphics[height=.8 in]{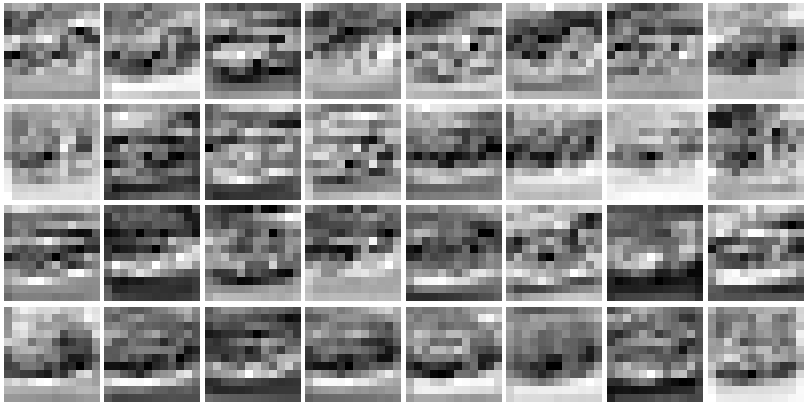}}
\subfigure[]{\includegraphics[height=.8 in]{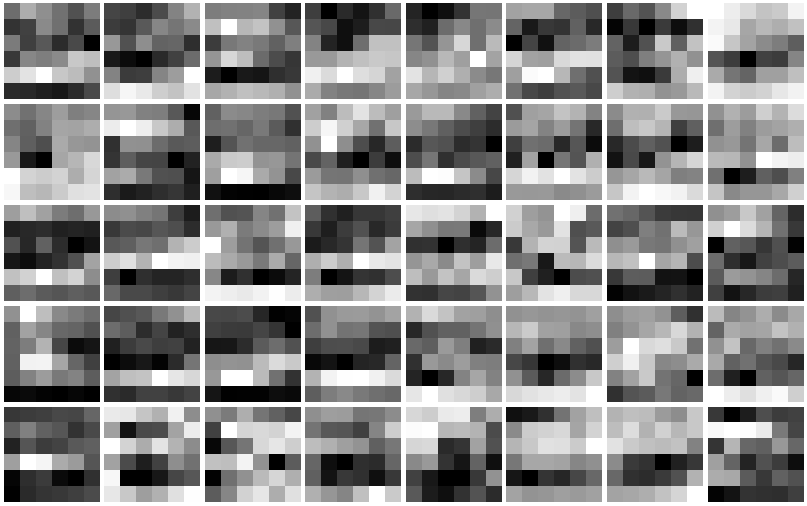}}
\subfigure[]{\includegraphics[height=.8 in]{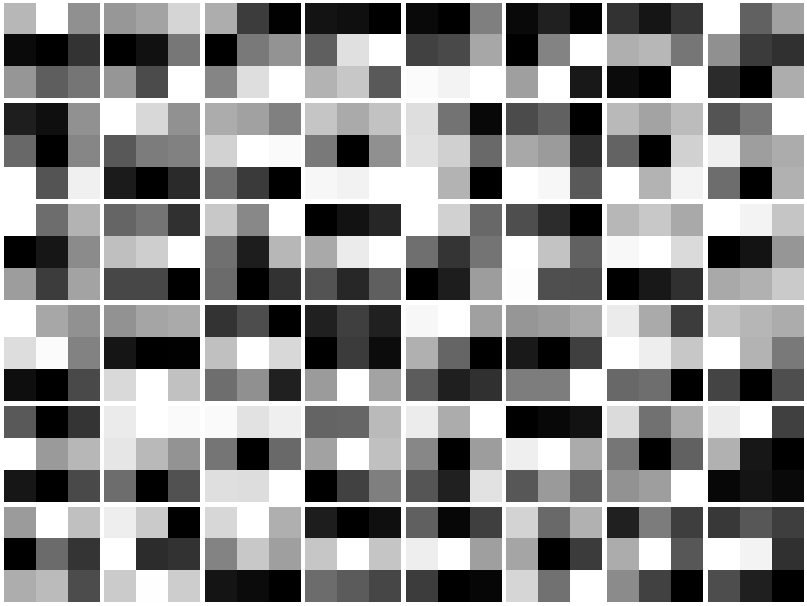}}

\caption{Visualizations of feature maps in ECN. The underlying network has $8$ feature maps in the first hidden layer. The feature channel growth rate is set to $8$. In this $6$-hidden-layer network there are $8,16,24,32,40,48$ feature maps in layers 1-6 respectively. In each deeper layer the newly generated higher level feature maps are appended as a new row. (a) Level 1 feature maps in the first hidden layer. (b) Level 1 features are adapted and tiled in the first row, level 2 features are newly generated using level 1 features, and are appended in the second row. (c) Both level 1 and level 2 features are adapted and tiled in the first two rows. Level 3 features are newly generated from level 1 and level 2 features, and are appended as the third row. (d) Level 4 features. (e) Level 5 features. (f) Level 6 features are used for recognition. The feature maps have been rescaled for better visualization.}
\label{fig:Maps}
\end{figure}
ECN is structured around the  insights that 1) maintaining low-level features through to the upper layers of a network is beneficial, 2) allowing multiple levels of features to interact with each other within the network is beneficial, 3) abrupt changes in feature map dimension could be less ideal than gradual changes, and 4) the manner in which features are conventionally combined within network blocks hampers the preservation of low-level features. Our architecture instantiates the first two of these insights through a ``cascade'' architecture - a two stream architecture, one stream for low-level features, one stream for successively higher level features, where these two streams interact at every layer. This differs from the existing method of using skip connections to introduce low-level features into upper level network layers in that low level features are maintained, and modulated appropriately, rather than simply jumping layers~\cite{Huang2016Densely}. This approach differs from a conventional two-stream approach~\cite{WaveletPackets} in that both streams interact with each other. For the third insight, in both streams of our cascade architecture bilinear interpolation is employed for fractional pooling, allowing a gradual rather than abrupt decrease of feature map dimensions. Different from existing architectures, in ECN a scaling factor is introduced which simplifies the design of the shape of the network removing the need for handcrafting network shape. For the fourth insight we remove the conventional combination of features across blocks in neural networks in order to preserve the low-level signal in ECN's two-stream cascade architecture. We demonstrate that preserving multilevel features enhances training by providing easy-to-train shortcuts.

Our evaluation of ECN resulted in several intriguing results: 1) With ECN's principled structuring, shallow networks (Fig. \ref{fig:CasNet}) seem to perform competitively well when compared to extremely deep networks. 2) Complex high level tasks such as image classification can be approached through evenly downsampling and adapting a set of highly structured features (Fig. \ref{fig:Maps}). 3) Low level features may be of critical importance in high level tasks such as image classification: not only do these features remain similar in the deeper adaptation process, but they may have provided major convenience for the training of high level features.

Finally, we evaluate ECN using multiple convolution block designs and find that recurrent and recursive designs lead to improved efficiency and accuracy. Without additional treatment, a standard convolution block design combined with recurrent connections leads to state-of-the-art accuracy in benchmark image classification tasks. Another block design using a recursive filter gives rise to state-of-the-art efficiency. Our 6-cascading-layer design with under 500k parameters achieves 95.24\% and 78.99\% accuracy on CIFAR-10 and CIFAR-100 datasets, respectively, outperforming the current state-of-the-art on small parameter networks, and our 3 million parameter version is competitive to the state-of-the-art.

\section{Related Work} \label{sec:related_work}

Even though convolutional networks have led to many exciting breakthroughs in visual and language learning tasks \cite{krizhevsky2012imagenet}, the mathematical understanding of convolutional networks is severely underdeveloped. It is intuitively clear that convolutional networks can be understood in part within the context of scale-space theory or multiresolution analysis \cite{Mallat:2008:WTS:1525499}. While an excellent attempt has been made \cite{UnderstandingCNN_Mallat} to explain convolutional networks in wavelet terms, it remains unclear how the techniques in wavelet theory can be explicitly used to simplify the construction and training of convolutional networks.

Skip connections or identity maps in the networks \cite{hochreiter1997long,he2016deep,Huang2016Densely,DPN} are a popular method for improving network performance. One intuitive understanding of the function of skip connections is that they pass lower level representations to deeper levels. In LSTM \cite{hochreiter1997long} and ResNet~\cite{he2016deep} networks the representations are adapted during this process.  Whereas in DenseNet~\cite{Huang2016Densely} shallow features are passed to the deeper levels without modification. One recent work, termed Dual Path Networks(DPN) \cite{DPN}, connects the two approaches using a high order recurrent neural network, and horizontally concatenates ResNet \cite{he2016deep} and DensNet \cite{Huang2016Densely}, and shows improved performance. 

Presently, the best performing networks are typically 50 to 100 layers deep~\cite{he2016deep,Huang2016Densely,DBLP:journals/corr/SrivastavaGS15}. Translating an existing backbone architecture to a new application with different input and output sizes is a nontrivial task. The structuring of these standard designs are constrained by integer resizing operations such as pooling and strided convolution. Fractional pooling and strides have been explored previously, e.g. \cite{graham2014fractional}. In our implementation of fractional pooling we employ bilinear interpolation, as is done in \cite{SpatialTransformer}. Furthermore, existing architectures merge different levels of features at the end of each block such that low- and high-level features become entangled and in-differentiable.  ECN safely removes these limitations.

As the networks become wider and deeper, methods such as pruning~\cite{DBLP:journals/corr/HanMD15,PruningHao,PruningICCV}, quantizing~\cite{DBLP:journals/corr/CourbariauxB16}, and knowledge distillation~\cite{DBLP:journals/corr/HintonVD15} have been introduced to reduce the size of these networks. ECN uses recursion\cite{RecurrentCNN,socher2012convolutional} to re-use parameters across layers, allowing for greater compactness of network design.
In the other direction, significant efforts have been spent on discovering more efficient network layers or overall architectures~\cite{DBLP:journals/corr/IandolaMAHDK16,MobileNet,DBLP:journals/corr/ZhangZLS17,Condensenet,Huang2016Densely,DBLP:journals/corr/ZhangQ0W17}.

\section{Methods} \label{sec:methods}

ECN is based on a simple cascading layer design. Intuitively, a cascading layer incorporates multilevel features and gradually resizes those features before providing them to the next layer. ECN consists of iteratively stacked cascading layers. This iterative construction process ends when a stopping condition is met - e.g., when feature map dimensions fall below a predefined size.

\subsection{Utilization of Multilevel Features}

\begin{figure}
\centering
\includegraphics[width=4.5 in]{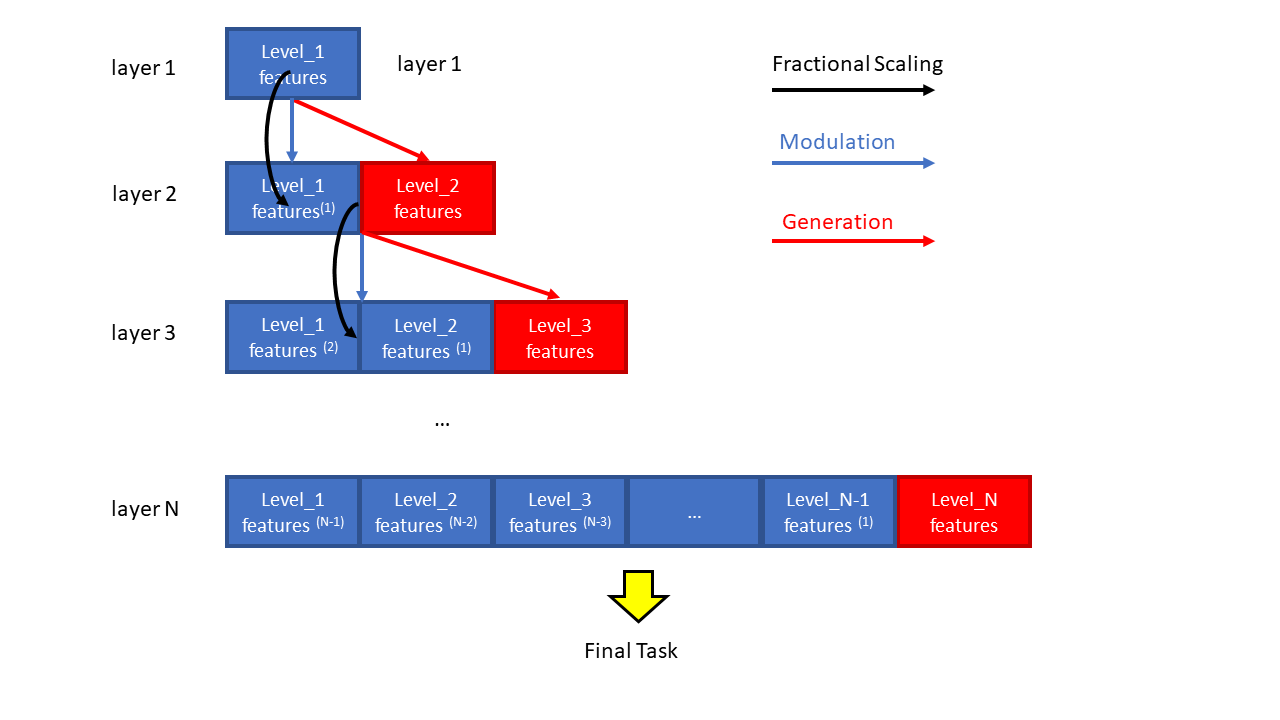}
\caption{Multilevel feature representations within ECN. For the $i$-th layer ($i$-th row in the figure), fractional scaling is used to downsample the features from layer $i-1$ (row $i-1$ in the figure). A convolution block is applied over these downsampled features to generate the modulation signal, and the level $i$ features (red block). The modulation signal is then added to the features downsampled from the  previous layer. The superscripts in each block represent the number of modulations the features have undergone.}
\label{fig:Multilevel}
\end{figure}

We now follow the intuition
that allowing an interplay between level of abstraction within perception can give rise to greater perceptual synergy 
  at the beginning of Section \ref{sec:introduction} in introducing ECN's usage of multilevel features. To illustrate ECN's relation to work in the literature we also provide three additional viewpoints of our utilization of multilevel features: one from the network architecture, one from wavelet analysis, and one from optimization of neural networks.

\subsubsection{From a Network Architectural Point of View}

Our cascading design can be viewed as evolving from the designs of several well-performing networks~\cite{he2016deep,Huang2016Densely,DPN}. Our first design can be viewed as a ResNet-inspired DenseNet, or a DenseNet-inspired ResNet. We make an extension to DenseNet~\cite{Huang2016Densely} when passing low level features to deeper layers by allowing them to be modulated by higher level features. The modulation of low level features can be viewed as a \textit{feedback} mechanism, where high level knowledge is providing ``advanced viewpoints'' to improve the low level knowledge, effected through a link from the higher level stream to the lower level stream of the network. The modulation signal is generated using a convolution block, and is added to the original features as in residual learning~\cite{he2016deep}. The generation of high level features is also achieved using the convolution block - however, the results are appended to the existing channels (Figs.~\ref{fig:CasNet},~\ref{fig:Multilevel}).  Similar to our design, but more costly, Dual Path Network~\cite{DPN} concatenates ResNet and DenseNet layers. By not passing low-level features through a $1\times1$ convolution, such as is done in DPN, ECN better preserves low-level features, making them more directly available at higher layers.

As a consequence of this ECN is able to pass features throughout the \textit{whole network}, rather than only within each \textit{block}. With this design we explicitly enforce that the shallowest level features be preserved throughout the whole network, with adaptations made only if they improve the final task (Figs.~\ref{fig:CasNet},~\ref{fig:Maps}).

\subsubsection{From a Wavelet Point of View}
It is noteworthy that the cascading layer is a more general form of the cascade algorithm used in wavelet packet decomposition~\cite{WaveletPackets,Mallat:2008:WTS:1525499}, where previous level signals are decomposed into a low frequency branch and a high frequency branch, usually followed by downsampling by a factor of $2$. In wavelet packet decomposition, the original signal is decomposed into a binary tree. The difference between this and ECN is that in ECN the two siblings are merged before the next level of decomposition, deviating from a tree structure. The extensions we made in our cascading design derive from the adaptations and downsampling of features which we introduce later.  Due to the cascading design, the evolution of feature maps within ECN closely resembles the evolution of modulus maxima in scale-space theory~\cite{Mallat:2008:WTS:1525499}. This relation suggests a path for further mathematical investigation of neural networks.

\subsubsection{From an Optimization Point of View}



ECN's construction facilitates training by providing easily trainable shortcuts to the optimization: we speculate that as a consequence of the two stream architecture, ECN allows the decomposition of $f$, the mapping from network input to output, into a series of progressively hierarchically deeper, and therefore harder to train, functions: $f=f_1+f_2+...+f_N$. See figure \ref{fig:Multilevel} for a visualization of function decomposition within and across layers. In figure \ref{fig:Multilevel} layer N contains features of levels $1$ through $N$ - this provides a direct link from the training (gradient) signal not only to higher level features, as is the case in other architectures, but to mid- and low-level features as well. This relaxes the interdependence in training between $f_i$ (low $f_i$ can be trained with less dependence on high $f_i$), and has the potential to facilitate training by allowing a training of $f$ component-wise. The result of this is the capability to train a complex function, $f$, by training the simpler functions of which it is composed, $f_i$.

Decompositions analogous to this kind have been proven to be helpful in multiscale signal analysis. Complex operations are simplified by conducting them at multiple scales, either (1) from coarse-to-fine or (2) in parallel. Here, we take the second approach to train different levels in parallel: the easier to train, or coarse, components can serve as a backbone model for the harder components, where the harder-to-train layers have only to compensate for the residuals of the learning problem.  We conjecture that ECN's construction facilitates the progressive training of neural networks. Note that ECN's construction is a revival of the layer-wise pretraining technique~\cite{Layer-wise-training} which triggered the era of deep learning. This classic approach followed the correct intuition of progressively developing high level representations. But the layer-wise training, which belongs to the first approach, lacks proper supervision signal.

\subsection{Fractional Scaling}

To facilitate the passing of low level features throughout the whole network, and to evenly resize the feature maps in a network, we propose to use fractional scaling, performed using bilinear interpolation, replacing the classic integer pooling and striding operations. Bilinear interpolation was first introduced in Spatial Transformer Networks~\cite{SpatialTransformer} to continuously deform feature maps for recognition tasks. Here we use bilinear interpolation to replace all the size change operations in the network.

Using bilinear interpolation, the outputs of the previous layer are fractionally scaled to the desired shape, and then serve as inputs to the next layer. Since bilinear interpolation is a locally smooth operation, subgradients can be calculated for backpropagation. In a network with a decreasing feature map dimension, having a constant scaling factor close to $1$ leads to a deep network. Similarly, having a constant scaling factor close to $0$ leads to a shallow network.

We adopt a simple uniform sampling when scaling the feature maps for obtaining the sample grid for bilinear interpolation. Non-uniform sampling is a natural extension~\cite{DeformableCNN} and we leave it for future work. Here, the output feature map dimension can be either calculated from the scaling factor or manually specified. The uniformly spaced sampling grid is then calculated to sample from the previous feature map. 

\subsection{Convolution Block Design}

\begin{figure}
\centering
\subfigure[]{\includegraphics[height=.75 in]{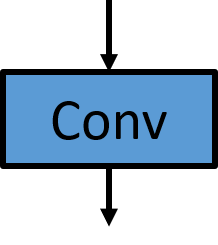}}
\subfigure[]{\includegraphics[height=.75 in]{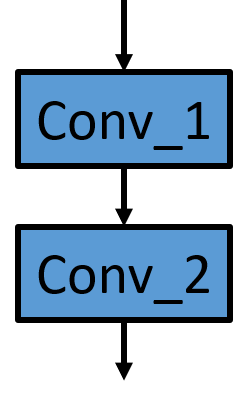}}
\subfigure[]{\includegraphics[height=.75 in]{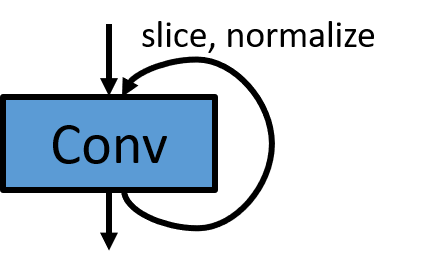}}
\subfigure[]{\includegraphics[height=.75 in]{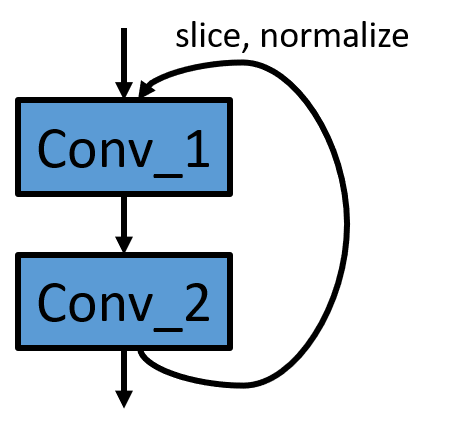}}
\subfigure[]{\includegraphics[height=.75 in]{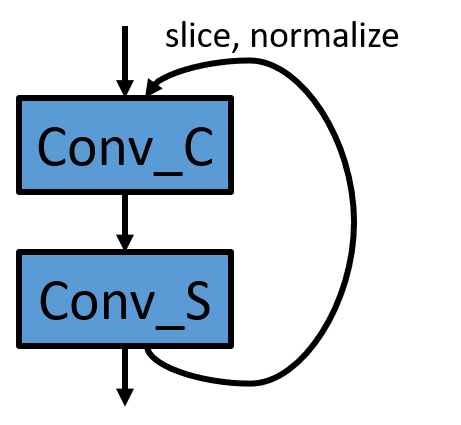}}
\subfigure[]{\includegraphics[height=.75 in]{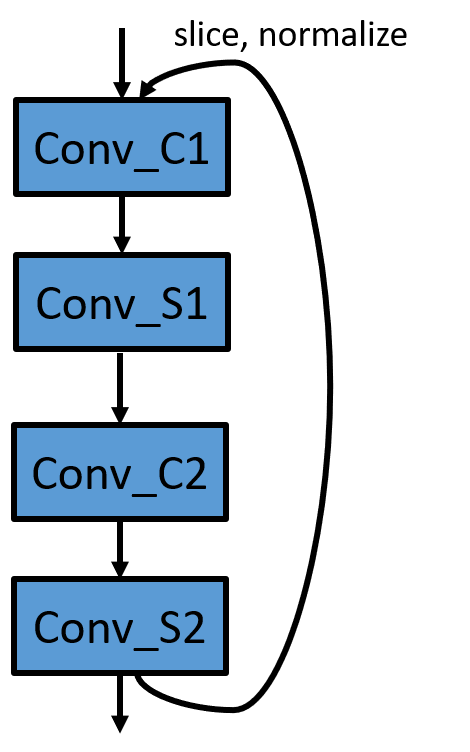}}

\caption{Convolution blocks that we evaluate: (a) Single convolution, (b) Double convolution, (c) Recurrent convolution, (d) Recurrent double convolution, (e) Recursive convolution. $Conv\_C$ stands for cross-channel $1\times 1$ convolution, $Conv\_S$ stands for channel-wise spatial convolution. (f) Recursive quadruple convolution.}
\label{fig:ConvBlocks}
\end{figure}

Many researchers have proposed different convolution block designs for use within layers of convolutional networks. However, in most cases different block designs are associated with different network architectures. As a consequence, it is difficult to draw conclusions about the relative merits of different block designs. However, with ECN's evenly cascaded design, it is straightforward to incorporate different block designs. In this work we propose and evaluate multiple block designs. More advanced block designs than are presented here can be evaluated in subsequent works and potentially result in improved performance. In this paper we include six basic blocks:

Block 1:  Single convolution (Fig.~\ref{fig:ConvBlocks}(a)). The convolutional layer has one single convolution operation, combined with standard techniques such as batch normalization~\cite{DBLP:journals/corr/IoffeS15} and the ReLU activation function~\cite{krizhevsky2012imagenet}. We order these operations as $BN \rightarrow ReLU \rightarrow Conv$. 

Block 2: Double convolution (Fig.~\ref{fig:ConvBlocks}(b)). Two single convolution layers are chained together, the number of intermediate output channels is set to be the larger of the input and output channels. 

Block 3: Recurrent convolution (Fig.~\ref{fig:ConvBlocks}(c)). We iteratively reuse the weights in Block 1 through recurrent connections. We made alterations to the previously reported approach~\cite{RecurrentCNN}. Firstly, the number of input channels do not usually match the number of output channels. Here we propose a simple solution: when the output channels ($OUT$) are more than the input channels ($IN$), we \textit{slice} the matching channels from the output (e.g. the first $IN$ channels) to reuse as input. Similar to an $LSTM$~\cite{hochreiter1997long} we add the outputs of the later iterations to the initial outputs. It is important to point out that our implementations of this recurrent design lead to worse results in ECN. The output signals usually have very different statistical properties from the input signals, and this inconsistency interfered with training. Our solution to this is that we insert an individual batch normalization operation for each iteration, leaving the convolution kernel weights shared in all iterations. This design significantly improves performance with very little increase in number of parameters (Tables~\ref{tab:CIFAR-10},~\ref{tab:CIFAR-100}, Fig.~\ref{fig:SizePerformance}).

Block 4: Recurrent double convolution (Fig. \ref{fig:ConvBlocks}(d)). Similar to Block 3 we iteratively reuse Block 2.

Block 5: Recursive convolution (Fig. \ref{fig:ConvBlocks}(e)). To make the convolutions more efficient we adopt separable convolution operations~\cite{MobileNet} to replace standard convolution. Cross-channel $1 \times 1$ convolution is applied in the first stage to change the number of input channels to match the number of output channels, followed by the channel-wise convolution in the second stage. Compared to standard convolution which involves more parameters, separable convolutions are usually more efficient but are weaker than directly applying standard convolutions. One fix for this weakness is to iteratively apply the filtering as in Blocks 3 and 4. Traditionally this technique is called recursive filtering. We adopt this name and include this class of filtering in our comparison. 

Block 6: Recursive quadruple convolution (Fig. \ref{fig:ConvBlocks}(f)). Each recurrent double convolution (Block 4) is replaced by four separable convolutions.

\section{Experiments} \label{sec:experiments}

\def \hfillx {\hspace*{-\textwidth} \hfill}

\begin{table}[t]
\begin{minipage}{0.7\textwidth}
\centering
\resizebox{\textwidth}{!}{
\begin{tabular}{@{}l|rrrrrr@{}}
\toprule
\textbf{Conv Block} & \textbf{\begin{tabular}[c]{@{}r@{}}Initial \\ Channels\end{tabular}} & \textbf{Scaling} & \textbf{CIFAR-10} & \textbf{Parameters} & \textbf{CIFAR-100} & \textbf{Parameters} \\ \midrule
type 4              & 16                                                                   & 3/4              & 93.49\%           & 214330              & 68.15\%            & 220180              \\
type 4              & 32                                                                   & 3/4              & 95.47\%           & 849130              & 76.05\%            & 860740              \\
type 4              & 64                                                                   & 3/4              & 96.20\%           & 3380170             & 79.80\%            & 3403300             \\
type 4              & 128                                                                  & 3/4              & 96.68\%           & 13488010            & 81.75\%            & 13534180            \\ \midrule
type 6              & 64                                                                   & 3/4              & 95.24\%           & 421770              & 78.99\%            & 444900              \\ \bottomrule
\end{tabular}
}
\caption{CIFAR accuracies for ECN-6}
\label{tab:Results}
\end{minipage}
\hfillx
\begin{minipage}{0.3\textwidth}
\centering
\resizebox{\textwidth}{!}{
\begin{tabular}{c}
\hline
\textbf{ECN-6}                                                                          \\ \hline
3x3 Conv                                                                        \\ \hline
\begin{tabular}[c]{@{}c@{}}
$\begin{bmatrix}Cascaded ConvBlock\\ Fractional Scaling
 \end{bmatrix}\times 6$
\end{tabular} \\ \hline
global average pooling                                                          \\ \hline
Softmax                                                                         \\ \hline  
\end{tabular}
}
\caption{ECN-6 architecture}
\label{tab:CasNet-6}
\end{minipage}
\end{table}

\subsection{Datasets}
We evaluate ECN over three datasets: CIFAR-10, CIFAR-100~\cite{krizhevsky2009learning}, and ImageNet-32~\cite{DBLP:journals/corr/ChrabaszczLH17}. These datasets consist of 32x32 pixel images. CIFAR-10 has 10 classes, CIFAR-100 has 100 classes, and ImageNet-32 has 1000 classes. We employ standard methods of data augmentation, including horizontal image flips, and random 32x32 crops of zero padded images, with 4 pixel padding. CIFAR-10 and CIFAR-100 each contains 50,000 training samples, and 10,000 testing samples. ImageNet-32 contains images of ImageNet~\cite{ImageNet}, downsampled to 32x32 pixels; it contains 1.2 million training samples, and 50 thousand validation samples.

\subsection{Results}


The ECN network we use has a shape where feature map dimensions consistently decrease in size; it is constructed by iteratively stacking cascading layers until the feature map size is below a preset threshold (4 pixels). In our experiments we fix the number of iterations in Blocks 3, 4, 5, and 6 to $3$.

In table \ref{tab:Results} we employed a scaling factor of $\frac{3}{4}$, resulting in an evenly cascaded structure with 6 cascading layers (ECN-6). We report results for differently scaled structures using block 4, and one more result using a more efficient design using block 6. At the beginning of the network, a convolution is used to transform the channel count of the input to $init\_channels$, which takes the value 16, 32, 64, or 128. In each consecutive layer, we generate channels of high level features with a growth rate of $init\_channels \times 2\times (1-scaling\_factor)$, corresponding to 8, 16, 32, 64 respectively in the four networks. The overall network architecture can be found in table \ref{tab:CasNet-6}. Here a cascaded convolution block~\cite{Mallat:2008:WTS:1525499} represents using one convolution block and grouping the results into a low level branch and a high level branch. Global average pooling is used to convert the final feature map into a vector for classification. To avoid overfitting we also insert dropout after ReLU activations for the 3 largest networks in table \ref{tab:Results}. The dropout rates range from 0.03-0.25. We use stochastic gradient descent to train the network for 2000 epochs with a batch size of 512, for CIFAR10 and CIFAR100, and for 50 epochs and a batch size of 512 for ImageNet-32. The training is scheduled with an initial learning rate of 0.1 and followed by cosine annealing learning rates. The results can be found in table \ref{tab:Results}.

\begin{table}
\begin{center}
\begin{tabular}{l|r|r|r}
\hline
\textbf{ImageNet32}  & \multicolumn{1}{c|}{\textbf{Params}} & \multicolumn{1}{c|}{\textbf{Top-1 error}} & \multicolumn{1}{c}{\textbf{Top-5 error}} \\ \hline
WRN-28-1 \cite{WideResNet}             & 0.44M                                & 67.97\%                                   & 42.49\%                                  \\
WRN-28-2 \cite{WideResNet}            & 1.6M                                 & 56.92\%                                   & 30.92\%                                  \\
WRN-28-5 \cite{WideResNet}            & 9.5M                                 & 45.36\%                                   & 21.36\%                                  \\
WRN-28-10 \cite{WideResNet}            & 37.1M                                & 40.96\%                                   & 18.87\%                                  \\ \hline
ECN-6, block 6 (32)  & 0.24M                                & 63.91\%                                   & 38.50\%                                  \\
ECN-3, block 6 (64)  & 0.48M                                 & 59.05\%                                   & 33.68\%                                  \\
ECN-6, block 6 (64)  & 0.68M                                & 55.51\%                                   & 30.26\%                                  \\
ECN-6, block 6 (128) & 2.1M                                 & 46.29\%                                   & 21.92\%                                  \\
ECN-3, block 4 (128) & 7.8M                                 & 45.10\%                                   & 21.06\%                                  \\
ECN-6, block 4 (128) & 14.0M                                & 41.87\%                                   & 18.61\%                                  \\ \hline
\end{tabular}
\end{center}
\caption{Results on the ImageNet-32 dataset, for WideResNet \cite{WideResNet}, and ECN (channel count)}
\label{tab:ImageNet32}
\end{table}

Our studies on ImageNet-32 demonstrate ECN's potential to generalize to larger scale datasets. We evaluate ECN-3 and ECN-6, corresponding to scaling rates of $\frac{1}{2}$ and $\frac{3}{4}$, over ImageNet-32. We use initial channel counts of 32, 64, and 128, and report results in table \ref{tab:ImageNet32}, with channel counts given in parentheses. ECN-6 is more efficient than the strong baseline results reported in WideResNet~\cite{WideResNet}. An ECN-6 network using only 2 million parameters shows comparable results to a WideResNet architecture using 9.5 million parameters. ECN with block 4 produces competitive results using smaller number of parameters than WideResNet. ECN-3, which has 3 cascading layers and 7 million parameters outperforms the 9.5 million parameter WRN-28-5 result. A larger ECN-6 network using block 4 with 14 million parameters achieves accuracy that is comparable to WRN-28-10, which contains 37.1 million parameters.

\begin{table}[t]
\begin{minipage}{0.49\textwidth}
\centering
\resizebox{\textwidth}{!}{
\begin{tabular}{l|r|r|r}
\hline
\textbf{Model}      & \multicolumn{1}{c|}{\textbf{Params}} & \multicolumn{1}{c|}{\textbf{CIFAR-10}} & \multicolumn{1}{c}{\textbf{CIFAR-100}} \\ \hline
VGG-16 pruned\cite{PruningHao}       & 5.4M                                 & 6.60                                   & 25.28                                  \\
VGG-19 pruned\cite{PruningICCV}       & 2.3M                                 & 6.20                                   & -                                      \\
VGG-19 pruned\cite{PruningICCV}       & 5M                                   & -                                      & 26.52                                  \\
Resnet-56 pruned\cite{PruningHao}    & .73M                                 & 6.94                                   & -                                      \\
Resnet-110 pruned\cite{PruningHao}   & 1.68M                                & 6.45                                   & -                                      \\
Resnet 164-B pruned\cite{PruningICCV} & 1.21M                                & 5.27                                   & 23.91                                  \\
DenseNet-40-pruned\cite{PruningICCV}  & .66M                                 & 5.19                                   & 25.28                                  \\
CondenseNet-94\cite{Condensenet}      & .33M                                 & 5.00                                   & 24.08                                  \\
CondenseNet-86 \cite{Condensenet}     & .52M                                 & 5.00                                   & 23.64                                  \\ \hline
ECN, Block 6   & .42M                                 & 4.76                                   & 21.01                                  \\ \hline
\end{tabular}
}
\caption{Error rate comparison with state-of-the-art efficient architectures}
\label{tab:EfficientArchitectures}
\end{minipage}
\hfillx
\begin{minipage}{0.49\textwidth}
\centering
\resizebox{\textwidth}{!}{
\begin{tabular}{lrrrlll}
\cline{1-4}
\multicolumn{1}{l|}{\textbf{Model}}        & \multicolumn{1}{r|}{\textbf{Params}} & \multicolumn{1}{r|}{\textbf{CIFAR-10}} & \textbf{CIFAR-100}   \\ \hline
\multicolumn{1}{l|}{ResNet-1001\cite{DBLP:journals/corr/HeZR016}}           & \multicolumn{1}{r|}{16.1M}           & \multicolumn{1}{r|}{4.62}              & 22.71                \\
\multicolumn{1}{l|}{Stochastic-Depth-1202\cite{DBLP:journals/corr/HuangSLSW16}} & \multicolumn{1}{r|}{19.4M}           & \multicolumn{1}{r|}{4.91}              & -                    \\
\multicolumn{1}{l|}{Wide-ResNet-28\cite{WideResNet}}        & \multicolumn{1}{r|}{36.5M}           & \multicolumn{1}{r|}{4}                 & 19.25                \\
\multicolumn{1}{l|}{ResNeXt-29\cite{DBLP:journals/corr/XieGDTH16}}            & \multicolumn{1}{r|}{68.1M}           & \multicolumn{1}{r|}{3.58}              & 19.25                \\
\multicolumn{1}{l|}{DenseNet-BC-190\cite{Huang2016Densely}}          & \multicolumn{1}{r|}{25.6M}           & \multicolumn{1}{r|}{3.46}              & 17.18                \\
\multicolumn{1}{l|}{NASNet-A*\cite{Condensenet}}             & \multicolumn{1}{r|}{3.3M}            & \multicolumn{1}{r|}{3.41}              & -                    \\
\multicolumn{1}{l|}{CondenseNet*light-160\cite{Condensenet}} & \multicolumn{1}{r|}{3.1M}            & \multicolumn{1}{r|}{3.46}              & 17.55                \\
\multicolumn{1}{l|}{CondenseNet-182\cite{Condensenet}}       & \multicolumn{1}{r|}{4.2M}            & \multicolumn{1}{r|}{3.76}              & 18.47                \\ \hline
\multicolumn{1}{l|}{ECN, Block4}         & \multicolumn{1}{r|}{3.3M}            & \multicolumn{1}{r|}{3.8}               & 20.2                 \\ \hline
\multicolumn{1}{l|}{ECN, Block4}         & \multicolumn{1}{r|}{13.3M}           & \multicolumn{1}{r|}{3.32}              & 18.25                \\ \hline
                                           & \multicolumn{1}{l}{}                 & \multicolumn{1}{l}{}                   & \multicolumn{1}{l}{}
\end{tabular}

}
\caption{Error rate comparison with state-of-the-art architectures}
\label{State of art architecture}
\end{minipage}
\end{table}

\subsection{Comparison of Convolution Blocks over CIFAR10 and CIFAR100}

We have compared the six block designs over CIFAR10 and CIFAR100 using various sized networks. The networks are trained for 500 epochs.
We tested scaling factors $\frac{1}{2}, \frac{3}{4},$ and $\frac{7}{8}$, and the corresponding networks have $3, 6,$ and $12$ cascading layers. The growth rates are calculated using the same strategy as explained above. For these experiments we use 3-stage learning rate scheduling, decreasing the learning rate at $40\%$ and $80\%$ total epoch count by a factor of 10. We set batch size to 512 for CIFAR-10. For CIFAR-100 a batch size of 128 usually leads to better performance, and we report the better of size 128 and size 512 batches. The results over CIFAR-10 and CIFAR-100 can be found in tables \ref{tab:CIFAR-10} and \ref{tab:CIFAR-100}, respectively.

\begin{figure}
\centering
\subfigure[] {\includegraphics[height=1.75in]{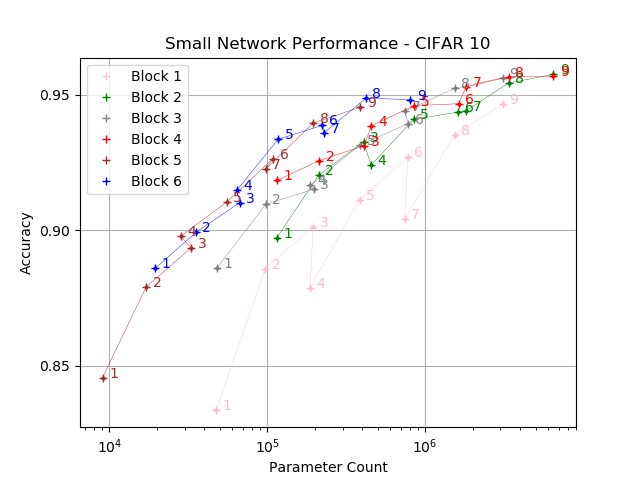}}
\subfigure[]{\includegraphics[height=1.75in]{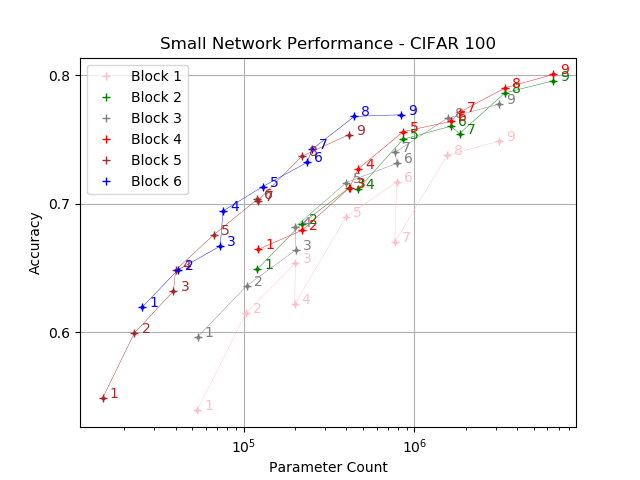}}
\caption{Relationship between network parameter count, block category, and classification accuracy for (a) CIFAR-10 and (b) CIFAR-100 datasets. The results are labeled according to the rows in which they appear in tables~\ref{tab:CIFAR-10} and ~\ref{tab:CIFAR-100}.}
\label{fig:SizePerformance}
\end{figure}

By comparing Blocks 1 and Blocks 3 in tables~\ref{tab:CIFAR-10} and \ref{tab:CIFAR-100}, and Fig.~\ref{fig:SizePerformance}, we found that reusing the convolution weights via recurrent connections significantly improves performance, while maintaining a small network size. When the convolution block becomes powerful, and especially when the model gets large, the improvement due to recurrent connections becomes smaller (Blocks 2 vs Blocks 4). Still, we find a surprise here that through recurrent connections even the single convolution can perform competitively with the widely used double convolution, using the same number of parameters (Fig.~\ref{fig:SizePerformance}). The optimal balance between depth and width varies from block to block. Our most efficient convolution block is block 6, which uses recursive quadruple convolutions. We reach state-of-the-art efficiency and the best results are reported in the last row of table~\ref{tab:Results}. It is noteworthy that although using separable convolution~\cite{MobileNet} reduces the number of parameters, the gain in efficiency also comes with a decrease in accuracy. The effective reduction in parameters enabled by using separable convolutions in ECN blocks $5$ and $6$ is around 2 fold to 4 fold. 

\begin{figure}[ht!]
\centering
\includegraphics[height=2.875 in]{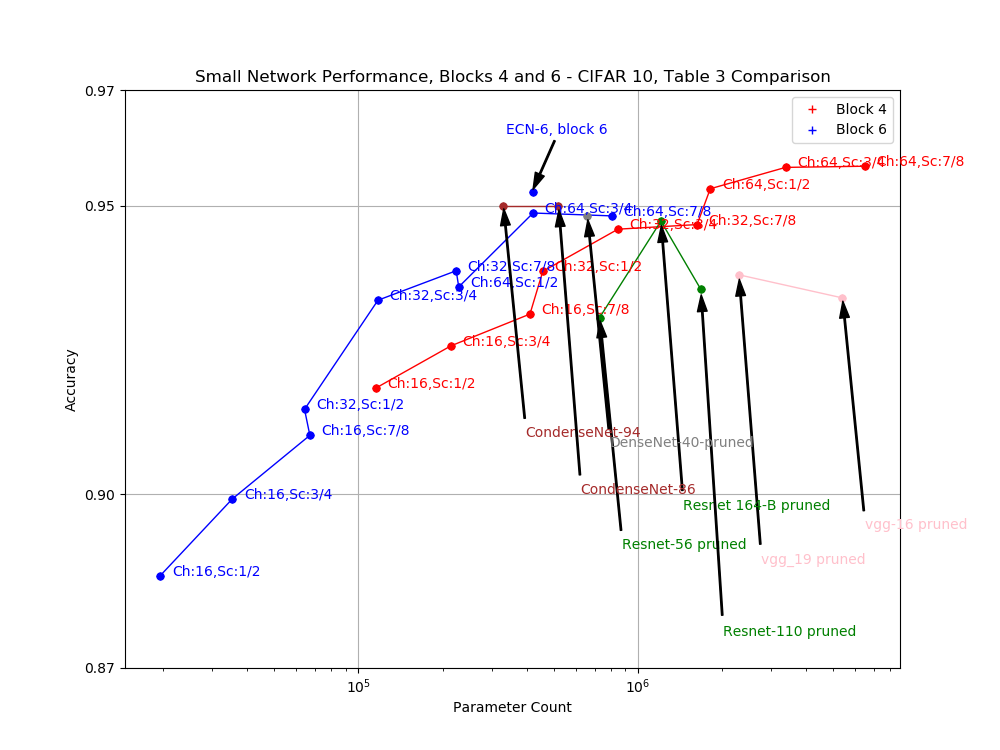}
\caption{Comparison between parameter count and classification accuracy over CIFAR-10 for ECN with blocks 4 and 6 and architectures listed in table ~\ref{tab:EfficientArchitectures}. Additionally, our best result using ECN-6 with block 6, attained through longer training, is plotted as ``ECN-6, block 6''.}
\label{fig:sizeperformancecifar10_4_6}
\end{figure}

\begin{figure}[ht!]
\centering
\includegraphics[height=2.875 in]{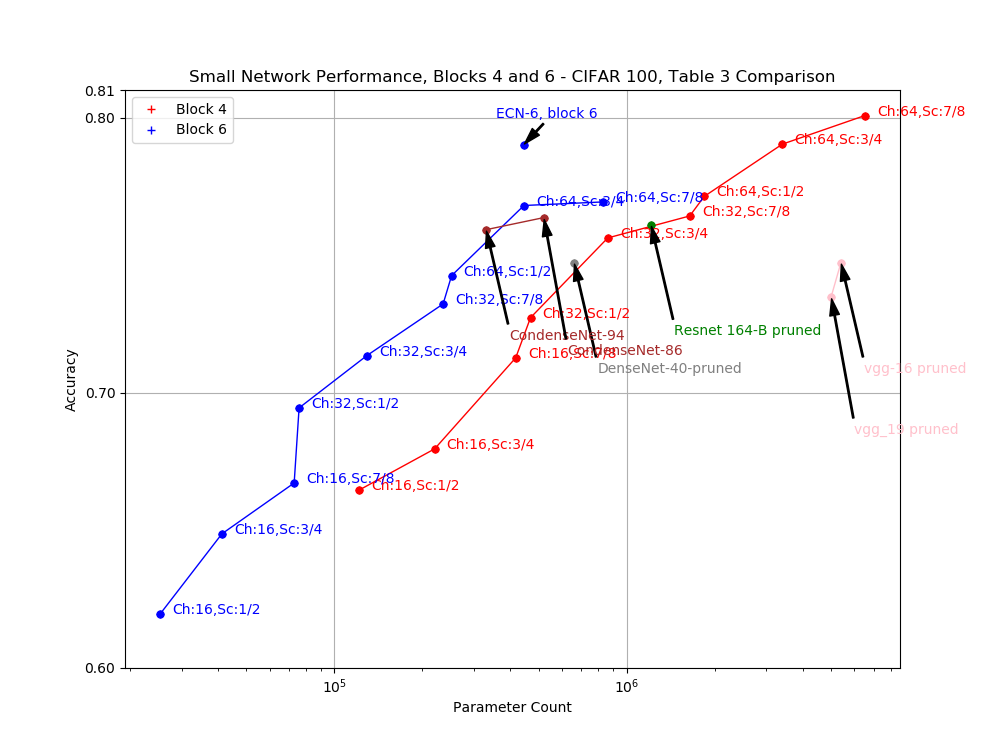}
\caption{Comparison between parameter count and classification accuracy over CIFAR-100 for ECN with blocks 4 and 6 and architectures listed in table ~\ref{tab:EfficientArchitectures}. Additionally, our best result using ECN-6 with block 6, attained through longer training, is plotted as ``ECN-6, block 6''.}
\label{fig:sizeperformancecifar100_4_6}
\end{figure}

When compared to other state-of-the-art efficient architecture designs, listed in table~\ref{tab:EfficientArchitectures}, ECN using block 6 achieves the lowest error rate without using any pruning methods. This is significant, as a simple and principled architecture design is proving to be better than sophisticated methods such as pruning described in~\cite{PruningICCV},~\cite{PruningHao} and even better than \cite{Condensenet} with smaller parameter count (Figs.~\ref{fig:sizeperformancecifar10_4_6}, ~\ref{fig:sizeperformancecifar100_4_6}). On the other hand, ECN block 4 does relatively well compared to other architectures listed in table~\ref{State of art architecture} that are using more advanced designs than ours.  

We have shown that there are avenues for improving the performance of convolutional networks by using principled designs like ECN. Even the simplest designs can reach state-of-the-art performance. Due to limitations in space and computational resources, only the 6 basic block designs are evaluated. More advanced block designs can modularly replace our basic block designs and potentially produce even better numbers. 

\begin{table}
\centering

\resizebox{\textwidth}{!}{
\begin{tabular}{rr|rr|rr|rr|rr|rr|rr}
\hline
\multicolumn{2}{c}{\textbf{CIFAR-10}}                              & \multicolumn{2}{c|}{\textbf{Block 1}}                                   & \multicolumn{2}{c|}{\textbf{Block 2}}                                   & \multicolumn{2}{c|}{\textbf{Block 3}}                                   & \multicolumn{2}{c|}{\textbf{Block 4}}                                   & \multicolumn{2}{c|}{\textbf{Block 5}}                                   & \multicolumn{2}{c|}{\textbf{Block 6}}                                  \\ \hline
\multicolumn{1}{c}{\textbf{Ch}} & \multicolumn{1}{c|}{\textbf{Sc}} & \multicolumn{1}{c}{\textbf{Acc}} & \multicolumn{1}{c|}{\textbf{Params}} & \multicolumn{1}{c}{\textbf{Acc}} & \multicolumn{1}{c|}{\textbf{Params}} & \multicolumn{1}{c}{\textbf{Acc}} & \multicolumn{1}{c|}{\textbf{Params}} & \multicolumn{1}{c}{\textbf{Acc}} & \multicolumn{1}{c|}{\textbf{Params}} & \multicolumn{1}{c}{\textbf{Acc}} & \multicolumn{1}{c|}{\textbf{Params}} & \multicolumn{1}{c}{\textbf{Acc}} & \multicolumn{1}{c}{\textbf{Params}} \\ \hline
16                              & 1/2                              & 83.35\%                          & 47482                                & 89.71\%                          & 114586                               & 88.59\%                          & 47866                                & 91.84\%                          & 115546                               & 84.56\%                          & 9066                                 & 88.59\%                          & 19514                               \\
16                              & 3/4                              & 88.58\%                          & 97258                                & 92.04\%                          & 212410                               & 90.96\%                          & 98122                                & 92.57\%                          & 214330                               & 87.90\%                          & 17090                                & 89.92\%                          & 35370                               \\
16                              & 7/8                              & 90.12\%                          & 195082                               & 93.26\%                          & 407194                               & 91.51\%                          & 196906                               & 93.12\%                          & 411034                               & 89.35\%                          & 32946                                & 91.02\%                          & 66986                               \\ \hline
32                              & 1/2                              & 87.88\%                          & 187114                               & 92.40\%                          & 454954                               & 91.66\%                          & 187882                               & 93.86\%                          & 456874                               & 89.78\%                          & 28362                                & 91.47\%                          & 64106                               \\
32                              & 3/4                              & 91.13\%                          & 385738                               & 94.09\%                          & 845290                               & 93.15\%                          & 387466                               & 94.59\%                          & 849130                               & 91.04\%                          & 55418                                & 93.36\%                          & 117450                              \\
32                              & 7/8                              & 92.70\%                          & 776074                               & 94.36\%                          & 1622506                              & 93.93\%                          & 779722                               & 94.67\%                          & 1630186                              & 92.64\%                          & 108762                               & 93.87\%                          & 223754                              \\ \hline
64                              & 1/2                              & 90.41\%                          & 742858                               & 94.41\%                          & 1813066                              & 94.39\%                          & 744394                               & 95.29\%                          & 1816906                              & 92.25\%                          & 97674                                & 93.59\%                          & 228554                              \\
64                              & 3/4                              & 93.53\%                          & 1536394                              & 95.43\%                          & 3372490                              & 95.24\%                          & 1539850                              & 95.66\%                          & 3380170                              & 93.96\%                          & 195818                               & 94.87\%                          & 421770                              \\
64                              & 7/8                              & 94.65\%                          & 3095818                              & 95.75\%                          & 6477514                              & 95.63\%                          & 3103114                              & 95.68\%                          & 6492874                              & 94.54\%                          & 389034                               & 94.82\%                          & 806666                              \\ \hline
\end{tabular}
}
\caption{Comparison of convolution blocks on the CIFAR-10 dataset}
\label{tab:CIFAR-10}

\end{table}

\begin{table}
\centering
\resizebox{\textwidth}{!}{
\begin{tabular}{rr|rr|rr|rr|rr|rr|rr}
\hline
\multicolumn{2}{c}{\textbf{CIFAR-100}}                             & \multicolumn{2}{c|}{\textbf{Block 1}}                                   & \multicolumn{2}{c|}{\textbf{Block 2}}                                   & \multicolumn{2}{c|}{\textbf{Block 3}}                                   & \multicolumn{2}{c|}{\textbf{Block 4}}                                   & \multicolumn{2}{c|}{\textbf{Block 5}}                                   & \multicolumn{2}{c|}{\textbf{Block 6}}                                  \\ \hline
\multicolumn{1}{c}{\textbf{Ch}} & \multicolumn{1}{c|}{\textbf{Sc}} & \multicolumn{1}{c}{\textbf{Acc}} & \multicolumn{1}{c|}{\textbf{Params}} & \multicolumn{1}{c}{\textbf{Acc}} & \multicolumn{1}{c|}{\textbf{Params}} & \multicolumn{1}{c}{\textbf{Acc}} & \multicolumn{1}{c|}{\textbf{Params}} & \multicolumn{1}{c}{\textbf{Acc}} & \multicolumn{1}{c|}{\textbf{Params}} & \multicolumn{1}{c}{\textbf{Acc}} & \multicolumn{1}{c|}{\textbf{Params}} & \multicolumn{1}{c}{\textbf{Acc}} & \multicolumn{1}{c}{\textbf{Params}} \\ \hline
16                              & 1/2                              & 53.93\%                          & 53332                                & 64.93\%                          & 120436                               & 59.64\%                          & 53716                                & 66.45\%                          & 121396                               & 54.88\%                          & 14916                                & 61.95\%                          & 25364                               \\
16                              & 3/4                              & 61.50\%                          & 103108                               & 68.45\%                          & 218260                               & 63.62\%                          & 103972                               & 67.95\%                          & 220180                               & 59.96\%                          & 22940                                & 64.86\%                          & 41220                               \\
16                              & 7/8                              & 65.40\%                          & 200932                               & 71.23\%                          & 413044                               & 66.43\%                          & 202756                               & 71.26\%                          & 416884                               & 63.21\%                          & 38796                                & 66.71\%                          & 72836                               \\ \hline
32                              & 1/2                              & 62.17\%                          & 198724                               & 71.12\%                          & 466564                               & 68.19\%                          & 199492                               & 72.72\%                          & 468484                               & 64.85\%                          & 39972                                & 69.44\%                          & 75716                               \\
32                              & 3/4                              & 68.98\%                          & 397348                               & 75.02\%                          & 856900                               & 71.66\%                          & 399076                               & 75.62\%                          & 860740                               & 67.61\%                          & 67028                                & 71.34\%                          & 129060                              \\
32                              & 7/8                              & 71.72\%                          & 787684                               & 76.04\%                          & 1634116                              & 73.16\%                          & 791332                               & 76.42\%                          & 1641796                              & 70.36\%                          & 120372                               & 73.22\%                          & 235364                              \\ \hline
64                              & 1/2                              & 67.02\%                          & 765988                               & 75.47\%                          & 1836196                              & 74.00\%                          & 767524                               & 77.13\%                          & 1840036                              & 70.23\%                          & 120804                               & 74.25\%                          & 251684                              \\
64                              & 3/4                              & 73.79\%                          & 1559524                              & 78.64\%                          & 3395620                              & 76.67\%                          & 1562980                              & 79.03\%                          & 3403300                              & 73.70\%                          & 218948                               & 76.80\%                          & 444900                              \\
64                              & 7/8                              & 74.87\%                          & 3118948                              & 79.57\%                          & 6500644                              & 77.76\%                          & 3126244                              & 80.07\%                          & 6516004                              & 75.38\%                          & 412164                               & 76.93\%                          & 829796                              \\ \hline
\end{tabular}
}
\caption{Comparison of convolution blocks on the CIFAR-100 dataset}
\label{tab:CIFAR-100}

\end{table}

\section{Acknowledgement}
The support of ONR under grant award  N00014-17-1-2622 and the support of the National Science Foundation under grants SMA 1540916 and CNS 1544787 are greatly acknowledged.

\section{Conclusion} \label{sec:conclusion}

Taking inspiration from cascading methods in wavelet packet decomposition, we have developed Evenly Cascaded convolutional Networks (ECN) for image tasks. ECN differs from other networks in the use two interacting streams - a high-level feature stream and a low-level feature stream. ECN's two streams allow for the promulgation of low-level features throughout the entire network, as well as the modulation of those low-level features using advanced perspectives from high-level features. The explicit use of multilevel features not only leads to highly capable networks but provides shortcuts for the training process. Additionally, ECN is structured such that feature map dimensions decrease in a consistent manner, removing burdens of ad hoc architecture design, and potentially improving feature preservation and utility. We have evaluated ECN over CIFAR-10 and CIFAR-100, obtaining state-of-the-art performance, for both datasets, for small network settings; and over ImageNet-32 ECN obtains competitive results.

\bibliography{references.bib}
\bibliographystyle{plain}

\end{document}